\documentclass{article}
\usepackage{ijcai20}

\usepackage{times}
\usepackage{soul}
\usepackage{url}
\usepackage[hidelinks]{hyperref}
\usepackage[utf8]{inputenc}
\usepackage[small]{caption}
\usepackage{graphicx}
\usepackage{amsmath}
\usepackage{amsthm}
\usepackage{booktabs}
\usepackage{algorithm}
\usepackage{algorithmic}
\title{Normalization of Input-output Shared Embeddings in Text Generation Models}

\author{
	Jinyang Liu$^1$\footnote{Contact Author}\and
	Yujia Zhai$^1$\and
	Zizhong Chen$^1$\\
	\affiliations
	$^1$University of California, Riverside\\
	\emails
	\{jliu447, yzhai015\}@ucr.edu,
	chen@cs.ucr.edu,
}

\begin{document}

\maketitle

\begin{abstract}
  Neural Network based models have been state-of-the-art models for various Natural Language Processing tasks, however, the input and output dimension problem in the networks has still not been fully resolved, especially in text generation tasks (e.g. Machine Translation, Text Summarization), in which input and output both have huge sizes of vocabularies. Therefore, input-output embedding weight sharing has been introduced and adopted widely, which remains to be improved. Based on linear algebra and statistical theories, this paper locates the shortcoming of existed input-output embedding weight sharing method, then raises methods for improving input-output weight shared embedding, among which methods of normalization of embedding weight matrices show best performance. These methods are nearly computational cost-free, can get combined with other embedding techniques, and show good effectiveness when applied on state-of-the-art Neural Network models. For Transformer-big models, the normalization techniques can get at best 0.6 BLEU improvement compared to the original version of model on WMT'16 En-De dataset, and similar BLEU improvements on IWSLT 14' datasets. For DynamicConv models, 0.5 BLEU improvement can be attained on WMT'16 En-De dataset, and 0.41 BLEU improvement on IWSLT 14' De-En translation task is achieved.
\end{abstract}

\section{Introduction}

Deep Neural Networks have been general solutionships for various Natural Language Processing tasks. For example, Transformer-based models \cite{vaswani2017attention,ott2018scaling} have been state-of-the-art models for tasks such as Machine Translation, and huge pre-trained Neural Network models \cite{devlin2018bert,yang2019xlnet,lan2019albert,raffel2019exploring} have dominated multi-task learning task in Natural Language Processing.

Although the design of the inner parts of the networks may vary, each Neural Network model needs modules for mappings between natural language words/characters/tokens and numerical vectors, Which is called embeddings.
 
\begin{figure}
	\centering
	\includegraphics[width=8cm, height=5cm]{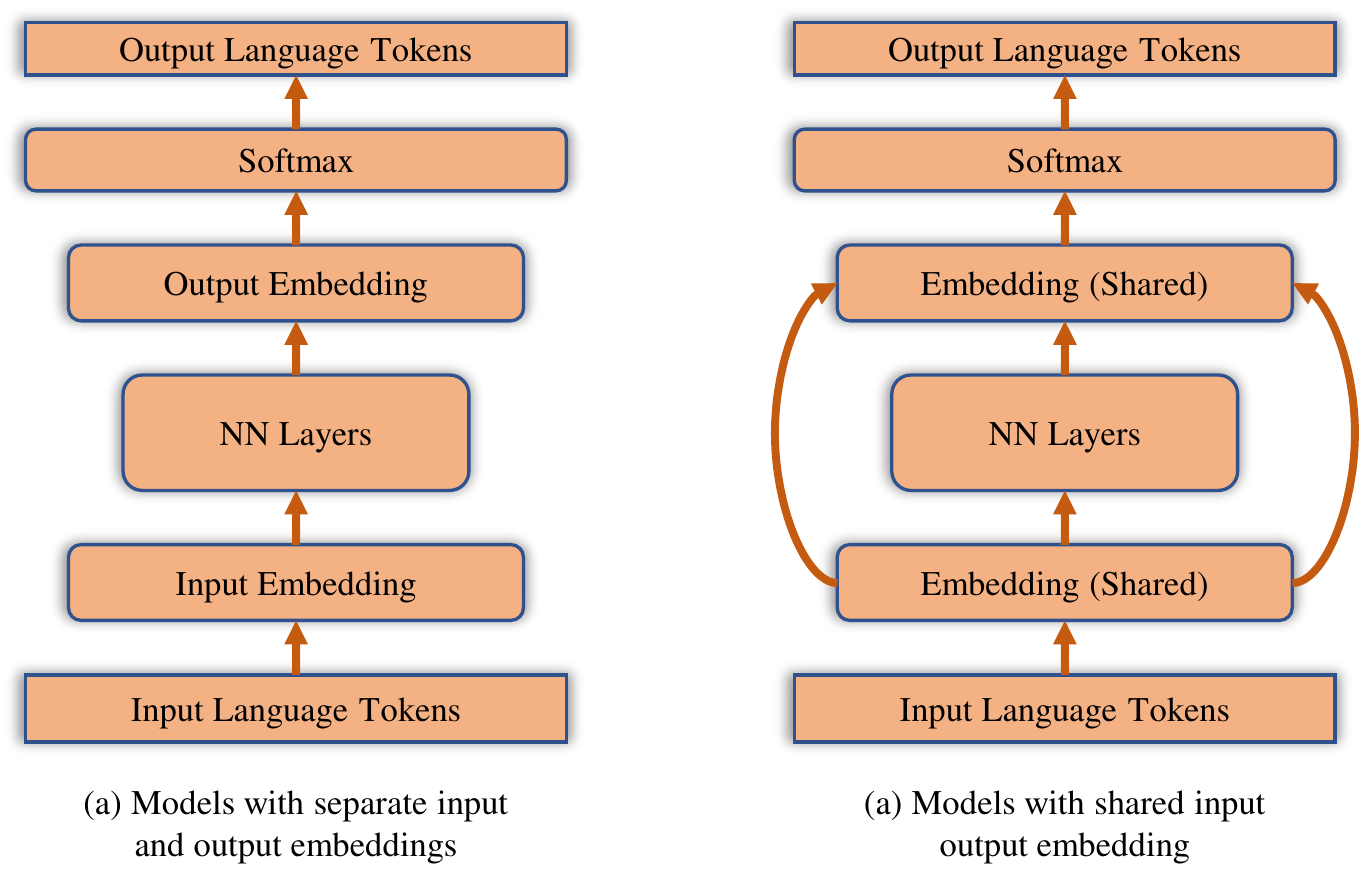}
	\caption{Neural network models with unshared and shared input-output shared embeddings. Model (a) on left has separate input and output embeddings, and model (b) on right shares embedding weights matrix through input and output layer.}
	\label{share}
\end{figure}

To relieve the great computational cost and storage need brought by huge size of embedding weights for input and output vocabularies, recent Neural Network models for text generation (e.g. Language Modeling or Neural Machine Translation models) are using input-output shared embedding weights \cite{press2016using} with a randomized initialization: When the input and output spaces are the same, the same embedding weights can serve in both input and output layers, like figure \ref{share} illustrates. The embedding weight sharing method greatly reduces the number of parameters of the model, at the same time preserves and even raises the performances of the model.

 Nevertheless, this paper points out that the weight-sharing method used in recent state-of-the-art Neural Network models has obvious shortcomings, and can be improved with low computational costs. After making insight into the existed input-output weight-sharing method, with linear algebra and statistical theories, this paper shows that normalization methods on the embedding weights should be applied on input-output shared embedding weights, and this kind of techniques can work well with shared embedding weight matrix in Neural Network text generation models. Moreover, the methods presented in this paper can also be well integrated with existed embedding parameter reduction method, such as Adaptive Embedding/Softmax \cite{grave2017efficient,baevski2018adaptive}, or Embedding Layer Factorization \cite{lan2019albert}.

 The normalization of word embedding have been discussed before, such as \cite{xing2015normalized}, but their works focus on different aspects from this paper (the domain-transfer problem, or density of word embedding). Totally speaking, the achievements of this paper are as following:

1. This paper makes deep analyis into and explains the effectiveness of input-output embedding weights sharing, then provides a theory-based method to improve it with trivial cost compared with the rest parts of the models.

2. The methods this paper presents can be simply integrated with other existed state-of-the-art embedding dimension reduction or parameter saving techniques.

3. In designed experiments, this paper proves the effectiveness of the methods in different kinds of Neural Network models for Natural Language Processing.

The outline of this paper is as following: Section 1 is the introduction of this paper, section 2 is the theory base and the practical normalization methods for input-output shared embeddings, section 3 is the experiment confurations, results and analysis. section 4 collects and introduces related works of this paper, and section 5 is a conclusion of this paper.

\section{Normalization of Input-output Shared Embedding}

\subsection{Decoder Output Embedding and Input-output Embedding Weights Sharing}

Input embedding and output Softmax layers are included in almost all Neural Network models for Natural Language Processing. Commonly speaking, An input embedding maps a language token (or regarded as an one-hot token vector) to a relatively low-dimension vector in continuous numerical space (mostly real Euclidean space).
Using $W$ to represent the input embedding Matrix, $\vec{x}$ to represent an one-hot input word vector (no batch size or sequence length is involved), the input embedding does the following computation:
\begin{equation}
\vec{e}=W\vec{x}
\end{equation}
 $W$ has the size of $D \times V$, $\vec{x}$ has the size of $V\times 1$ (all 1D vectors are as default column vectors in this paper) and the generated embedding vector $\vec{e}$ has size of $D \times 1$. $V$ is the vocabulary size which is often big (e.g. tens of thousands), and $D$ is the input embedding dimension.  
 
At the output layer, suppose $\vec{h}$ is the output embedding vector generated by the inner parts of the decoder, which is for an output token, the output Softmax layer does the following computations to find the most probable token:

\begin{equation}
scores=U\vec{h}+B
\end{equation}
\begin{equation}
probs=Softmax(scores)
\end{equation}
\begin{equation}
predtokenid=argmax(probs)
\end{equation}

$U$ is a $V\times H$ matrix for Softmax layer kernel, and $B$ is a vector with size $V$ for layer bias. $H$ is the size of the output embedding vector.

In some text generation Natural Language Processing tasks, the input and output token space is the same, e.g. language modeling, text summrization, and machine translation (when using shared vocabulary). In this situation, if the output embedding vector and the input embedding vector has the same size ($H=D$), it can be asserted that the space of output embedding vectors is isomorphy to the input embedding space. That is, given a output embedding vector $\vec{h}$, it should be a non-negative $l1$-normalized linear combination of input embedding vectors $\vec{w_i}$ (pay attention that $\vec{w_i}$ is the ith column of matrix $W$).
\begin{equation}
\vec{h}=\sum_{i}\alpha_i\vec{w_i}=W\vec{\alpha}
\end{equation}

In which $\vec{\alpha}=(\alpha_i)^T$ is a probability distribution (it's non-negative and normalized) showing the probabilites of the feature vector should generate the i-th token for output, which is just the conditional probabality $P(Y_i|X)$ of output and input. Apparently, Softmax layers can generate $\vec{\alpha}$, and a model can also use $scores$ vector to estimate $\vec{\alpha}$, as Softmax transformation has the isotone property. Because embedding dimension $D$ is always smaller than the vocabulary size $V$, the problem of solving $\vec{\alpha}$ is on an underdefined linear equations set. Therefore, additional constraints need to be introduced into this problem. Using a priori knowledge, one can confidently make the assumption that the real distribution of $\vec{\alpha}$ should be sparse (there will be little number of tokens which are appropriate for output at a certain position of one text sentence), and the whole optimization problem is:
\begin{alignat}{2}
\min  \vert\vert{\vec{\alpha}}\vert\vert_0\\
\mbox{s.t.} \quad \vec{\alpha}>=0,\\
 \quad \vert\vert{\vec{\alpha}}\vert\vert_1=1,\\
 \quad W\vec{\alpha}=\vec{h}.
\end{alignat} 
Or its relaxed convex version:
 \begin{alignat}{2}
 \min  \vert\vert{\vec{\alpha}}\vert\vert_1\\
 \mbox{s.t.} \quad \vec{\alpha}>=0,\\
 \quad \vert\vert{\vec{\alpha}}\vert\vert_2=1,\\
 \quad W\vec{\alpha}=\vec{h}.
 \end{alignat} 
 
But this convex version is still computational-expensive to be directly solved. As it is for an output layer of a big Neural Network, it is natural to hope that the estimation can be computed with simplified computations. Because Neural Network models just use the estimations for Softmax/Argmax classification, only the order of $\alpha_i$ (the components of $\vec{\alpha}$ on each dimension, which is the estimation score for each token) determines the output. Thus, one can also get isotone estimations $\hat{\alpha_i}$ for ${\alpha_i}$ instead, which have the property of:
\begin{equation}
\hat{\alpha_i}<\hat{\alpha_j} \quad if \quad \alpha_i<\alpha_j 
\end{equation}
And the estimations $\vec{\hat{\alpha}}$ are expected to be computed in a linear way, like the Fully Connected Layer:
\begin{equation}
\hat{\vec{\alpha}}=U^T\vec{h}+\vec{b}
\end{equation}
Where $U$ is a matrix having the same size of $W$ (it serves for the kernel of the Fully Connected Layer), and $\vec{b}$ is the bias vector of the Fully Connected Layer. The most direct method is learning $U$ and $\vec{b}$ from scratch. However, the size of $U$ is huge, so weight sharing methods attempt to bind $U$ and $W$ together with simple operations.
 So far, the most widely used, and almost only used setting for $U$ and $\vec{b}$ is $U=W$ , and $\vec{b}=0$ (it will be the baseline in this paper's experiments). It is introduced in \cite{press2016using}, and has achieved good performances, therefore has been adopted by many state-of-the-art Natural Language Processing models. In the next subsection, it will be discussed how to improve this method with simple and low cost computations.

\subsection{Normalization in Embedding Weight Matrix}
First, let's see specifically what $\hat{\vec{\alpha}}=U^T\vec{h}+\vec{b}$ the baseline method is using. Still assume that $\vec{h}$ as the real distribution of $\vec{h}=\sum_{i}\alpha_i\vec{w_i}=W\vec{\alpha}$, therefore,
\begin{equation}
\\hat{\vec{\alpha}}=U^T\vec{h}+\vec{b}=W^T\vec{h}=W^TW\vec{\alpha}
\end{equation}
\begin{equation}
\hat{\vec{\alpha}}=U^T\vec{h}+\vec{b}=\sum_{j}\vec{w_i}^T\vec{w_j}\alpha_j
\end{equation}
Remind that $\vec{w_i}$ is the i-th column of $W$ (the input embedding vector of the i-th token). As the target of Neural Network models is to find argmax, it is most important in the estimation of the largest $\alpha_i$. let $k=argmax_i\alpha_i$, then
\begin{equation}
\hat{\alpha_k}=\sum_{i}\vec{w_k}^T\vec{w_i}\alpha_i \approx \vec{w_k}^T\vec{w_k}\alpha_k=\vert\vert{\vec{w_k}}\vert\vert_2\alpha_k
\end{equation}
The approximation is based on the fact that $\vec{\alpha}$ is sparse, and $\alpha_k$ is much larger than others. Obviously, using the baseline method, the estimation of $\alpha_k$ is biased: For token with small embedding vector norm, it is under-estimated, and for token with big embedding vector, it is over-estimated. To solve this problem, this paper designs several improvement methods with small additional computation cost, and the first two are embedding normalization methods.

\subsubsection{$l_2$-normalized Input Embedding Weight Matrix}

The most direct way for getting free of the norm problem is to use a column-wise $l_2$-normalized input embedding matrix W having the structure of:
\begin{equation}
W=(\frac{\vec{w_1}}{\vert\vert\vec{w_1}\vert\vert_2}, \frac{\vec{w_2}}{\vert\vert\vec{w_1}\vert\vert_2}, ... \frac{\vec{w_V}}{\vert\vert\vec{w_V}\vert\vert_2})
\end{equation} 

And the $U$ still remain the same as $W$. Using $l_2$-normalized embedding Matrix will lead to a unbiased estimation of $\alpha$. The proof is as following: For $\vec{w_i}$ and $\vec{w_j}$ which are uniformly i.i.d on unit sphere, $\vec{w_i}^T\vec{w_i}=1$ and

\begin{equation}
E(\vec{w_i}^T\vec{w_j})=0
\end{equation}

Because due to symmetry it is easy to prove that the surface integral
\begin{equation}
\oint_{\sum_i{x_i^2}=1} a_ix_idx_1dx_2...dx_D=0
\end{equation}
when $\sum_i{a_i^2}=1$. Therefore, from equation (17), $E(\hat{\alpha_k})=\alpha_k$. The variance of $\hat{\alpha_k}$ will be small when $k$ is $argmax_i\alpha_i$.

A similar method is raised in \cite{nguyen2017improving} and validated in \cite{nguyen2019transformers}. One problem for this method is that it abandons the diversity of norms of embedding vectors, which also includes semantic information.
\subsubsection{Square-normalized Output Embedding}

If let $B=0$ and $U$ equals:
\begin{equation}
U=(\frac{\vec{w_1}}{\vert\vert\vec{w_1}\vert\vert_2^2}, \frac{\vec{w_2}}{\vert\vert\vec{w_2}\vert\vert_2^2}, ... \frac{\vec{w_V}}{\vert\vert\vec{w_V}\vert\vert_2^2})
\end{equation}
So that 
\begin{equation}
\hat{\alpha_k}=\sum_{i}\frac{\vec{w_k}^T\vec{w_i}\alpha_i}{\vert\vert\vec{w_k}\vert\vert_2^2} \approx \alpha_k 
\end{equation}
when $k$ is $argmax_i\alpha_i$. By scaling the output embedding kernel matrix by square norm, the estimation $\hat{\alpha_k}$ is also a unbiased estimation, and the value of $\alpha_k$ will never be over-estimated with large embedding vector norm. However, this method severely over-estimated the $\alpha_k$ with a small embedding vector norm, as this method do not consider the non-negative $l_1$-normalized constraints of $\vec{\alpha}$, which means that $\hat{\alpha_k}$ will get no punishment and is even encouraged for output when it is over 1.

For Comparison and proof for the effectiveness of Embedding Normalization, this paper introduces 2 more methods of sharing input and output embeddings. These 2 methods are similar to embedding normalization, but are in different principles.

\subsubsection{Calculate Distances for Estimation}
 Another way to find the $k$ for biggest $\alpha_k$ is to calculate the "similarities" between $\vec{h}$ and $\vec{w_i}$. A scale-sensitive method is to calculated the $l_2$-distance:
\begin{equation}
d_i=\vert\vert\vec{h}-\vec{w_i}\vert\vert_2
\end{equation}
For smoothness, the square of it is used:
\begin{equation}
d_i^2=(\vec{h}-\vec{w_i})^T(\vec{h}-\vec{w_i})=\vert\vert\vec{w_i}\vert\vert_2^2-2\vec{w_i}^T\vec{h}+\vert\vert\vec{h}\vert\vert_2^2
\end{equation}
As $\vert\vert\vec{h}\vert\vert_2^2$ is the same for each $i$, and we'd like to find the mininum of $d_i^2$, the $score_i$ can be in the following form:
\begin{equation}
score_i=\vec{w_i}^T\vec{h}-\frac{1}{2}\vert\vert\vec{w_i}\vert\vert_2^2
\end{equation}
and it can be implemented by just setting $U=W$ and $B=(\frac{1}{2}\vert\vert\vec{w_i}\vert\vert_2^2)^T$.

\subsubsection{Calculate Cosine Similarity for Estimation}

This method reviews the value of $\vec{h}$, and holds the view that the scale of $\vec{h}$ may be unimportant, as the signal generated by the Neural Network layers can be amplifiled or reduced, and the most vital information is in its direction. Therefore, this method computes cosine-similarities between $\vec{h}$ and $\vec{w_i}$:
\begin{equation}
U=(\frac{\vec{w_1}}{\vert\vert\vec{w_1}\vert\vert_2}, \frac{\vec{w_2}}{\vert\vert\vec{w_2}\vert\vert_2}, ... \frac{\vec{w_V}}{\vert\vert\vec{w_V}\vert\vert_2})
\end{equation}
\begin{equation}
score_i=\frac{\vec{w_i}^T\vec{h}}{\vert\vert\vec{w_i}\vert\vert_2}=cosinesim(\vec{h},\vec{w_i})*\vert\vert\vec{h}\vert\vert_2
\end{equation}
It relaxed the constraint of $\vert\vert\vec{\alpha}\vert\vert_1=1$ and attempts to found the k of:
\begin{equation}
k=argmin_k({min_a \vert\vert\vec{h}-a\vec{w_k}\vert\vert_2})
\end{equation}
If the $\vec{h}$ is scale-free, it will get better performances.

To theoretically evaluate these estimation methods, the following 3 important properties are raised, which a good estimation of $\alpha$ should have, but the original embedding weight sharing method does not preserve.

1. Identity: if $\vec{h}=\vec{w_k}$, then $k=argmax_i{\hat{\alpha_i}}$. This property means that, if the output embedding vector is the same as one input embedding vector, the output layer should return the corresponding token as output. The original embedding sharing method do not have this property. $l_2$-normalized Input Embedding Weight Matrix and Calculate Cosine Similarity for Estimation have this attribute, and the other 2 methods have this when there exists no $\vec{w_i}$ and $\vec{w_j}$ that $\vec{w_i}=a\vec{w_j}$ (in almost all practical circumstances it is true).

2. Normality: $\hat{\alpha_i} \le 1$, therefore $\hat{\alpha_i}$ always represents a probability. Only $l_2$-normalized Input Embedding Weight Matrix has this property.

3. Unbiased: $E(\hat{\alpha_i})=\alpha_i$. The 2 Embedding Normalization methods both have this property.

At last, the additional of computational costs of the 4 methods are all on computing the columnwise norms of $W$, which has the time complexity of $O(DV)$, and is the same as the original output Softmax Layer. Therefore, it is still trivial compared with the inner layers of Neural Network models.

In the following parts of the experiments, this paper will show the specific evaluation results and provide analysis for the 4 input-output embedding weight sharing methods.

\section{Experiments and Results on Machine Translation Task}
This section presents experiment settings, configurations ,results and analysis of the experiments this paper carries out for validating the effectiveness of embedding normalization methods. The experiments are based on Machine Translation task, because this task is one of the most important Text Generation tasks, and has clear benchmarks. The techinuqes can be also applied into other text generation tasks, as long as shared input-output shared embedding is included.

\subsection{Experimental Settings}

The embedding normalization methods designed by the paper are not constricted to one certain kind of models, but are compatible with various kinds Neural Network models with input-output shared embedding. This paper implements the methods on 2 kinds of state-of-the-art models in Neural Machine Translation task: Transformer \cite{vaswani2017attention} and DynamicConv \cite{wu2019pay} models on the following Datasets: WMT 16' En-De, and IWSLT 14' En-De, En-Es, and En-Ro datasets. The Transformer model is trained and tested on all datasets used, and the DynamicConv Models is trained and tested on WMT 16' En-De and IWSLT 14' En-De datasets.

\subsubsection{Datasets}

On WMT 16' En-De dataset, this paper follows the same settings from \cite{vaswani2017attention,ott2018scaling}, uses 4.5M sentence pairs for training, validates on newstest2013 and tests on newstest2014. The vocabulary has 32K symbols generated by sentencepiece \cite{kudo2018sentencepiece} and BPE \cite{sennrich2015BPE}.

On IWSLT 14' datasets, This paper uses English as target language, and uses German (De), Spanish (Es) and Romania (Ro) as source language. For the dataset splitting method of IWSLT 14' datasets, in De-En translation task. the paper follows the same settings from \cite{edunov2017classical} and \cite{xia2019tied}, and in Es-En and Ro-En tasks, this paper uses $IWSLT14.TED.dev2010$ as valid set and combines $IWSLT14.TED.tst2010/2011/2012$ for test set.

\subsubsection{Model Configurations}
For Transformer model on WMT'16 En-De dataset, this paper applies the Transformer-big configuration described in \cite{vaswani2017attention}, which has 6 blocks in both encoder and decoder parts, embedding dimension of 1024, hidden layer size of 4096, and 16 heads in multihead attention block. The label smoothing rate is 0.1. The training configurations are referenced from \cite{ott2018scaling}, this paper simulates an 128-GPU training process with half-precision and $458k$ batch size, 0.3 dropout rate, and an Adam optimizer with $\beta_1=0.9$, $\beta_2=0.98$, and $\epsilon=1e-8$. The learning rate scheduler is the same as \cite{vaswani2017attention}, with the summit learning rate of $1e-3$. 

For Transformer model on IWSLT 14' datasets, this paper applies a Transformer model which has 6 blocks in both encoder and decoder parts, embedding dimension of 512, hidden layer size of 1024, and 4 heads in multihead attention block. The label smoothing rate is 0.1. The training configurations are almost same as the Transformer in WMT dataset, except that a weight-decay rate of $1e-4$ is used. Default configurations for ISWLT 14' translation Transformer models use separate input and output vocabularies, but De-En translation model in this paper uses the same joined vocabulary like models on WMT 16' En-De dataset, because this paper finds that this way can improve BLEUs. Therefore, in IWSLT 14' De-En, all embedding weights are shared, and in Es-En and Ro-En only decoder embedding weights are shared.

For DynamicConv Models, as the model structures are similar between Transformer and DynamicConv models, the model configurations are also nearly the same between the models on corresponding datasets, except that DynamicConv models have 7 blocks in encoder parts. The training configuration is the same as \cite{wu2019pay}. 

All models are implemented with the fairseq-py toolkit \cite{fairseq} in PyTorch. The train and test are done on NVIDIA RTX 2080Ti, NVIDIA TESLA P100 and NVIDIA TESLA T4 GPUs.

\subsection{Results}
\subsubsection{Evaluation Method}
Like all other Neural Machine Translation experiments, this paper uses BLEU\footnote{https://github.com/moses-smt/mosesdecoder/blob/master/scripts/\\generic/multi-bleu.perl} as the evaluation method for the quality of generated translation texts.

The translation texts from trained models are generated with beam-search. For Transformer-big model on WMT 16' En-De dataset, texts are generated with beam width of 4 and length penalty of 0.6. For Transformer models on IWSLT 14' datasets, the beam width is set to 5 and the length penalty is set to 1. For DynamicConv model on WMT 16' En-De dataset, texts are generated with beam width of 5 and length penalty of 0.5. For DynamicConv models on IWSLT 14' datasets, the beam width is set to 4 and the length penalty is set to 1. For Transformer and DynamicConv models on WMT 16' En-De dataset, this paper makes average on the last 10 checkpoints for test.

\subsubsection{Experimental Results}

	Table \ref{TFWMT} and \ref{DCWMT} shows the BLEUs of Transformer-big models and DynamicConv models on WMT 16' En-De dataset with different embedding sharing configurations (Baseline is the original sharing method). The model can gain BLEU improvement from each of the 4 designed method, and using $l_2$-normalized input embedding get the best result for both models on WMT 16' En-De dataset. As analyzed before, embedding sharing method of $l_2$-normalized input embedding has good properties, that can give normalized and unbiased estimations of tokens' scores for output. For Transformer-big model, 0.6 BLEU improvement is achieved and for DynamicConv model, 0.5 BLEU improvement is achieved.
	
\begin{figure}[]
		\centering
		\includegraphics[width=8cm,height=6cm]{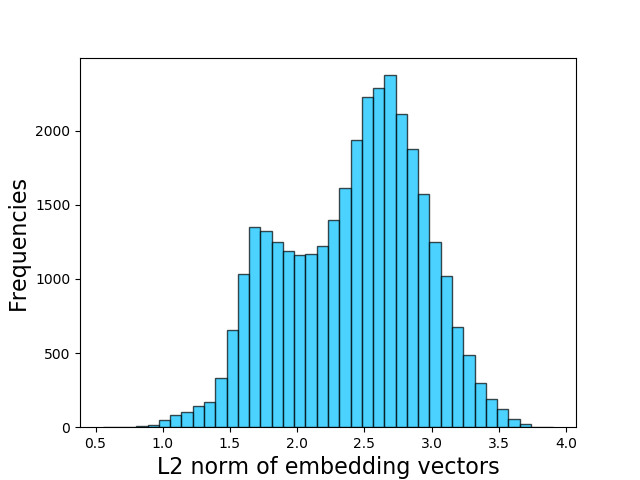}
		\caption{Frequency histogram of $l_2$ norms of embedding vectors}
		\label{histo}
	\end{figure}

Figure \ref{histo} is a histogram of $l_2$-norms of embedding vectors for language tokens in a baseline Transformer-big model on WMT 16' En-De Dataset. The histogram illustrates that the embedding vectors are scattered through training phase, get spreaded to have varied norm, which affects the quality of output scores with the original output embedding. 	
\begin{table}[]
	\centering
	\begin{tabular}{lrr}  
		\toprule
		Configuration & BLEU \\
		\midrule
		Baseline  & 29.3      \\
		$l_2$-normalized input embedding   & $\textbf{29.9}$       \\
		Square-normalized output embedding  & 29.4       \\
		Distances  & 29.8       \\
		Cosine similarities  & 29.7       \\
		\bottomrule
	\end{tabular}
	\caption{BLEUs of Transformer-big models on WMT 16' En-De dataset with different embedding sharing configurations.}
	\label{TFWMT}
\end{table}

\begin{table}[]
	\centering
	\begin{tabular}{lrr}  
		\toprule
		Configuration & BLEU \\
		\midrule
		
		Baseline  &    29.7   \\
		$l_2$-normalized input embedding  & $\textbf{30.2}$      \\
		Square-normalized output embedding  &  30.0      \\
		Distances  &  29.8     \\
		Cosine similarities  &  29.9      \\
		\bottomrule
	\end{tabular}
	\caption{BLEUs of DynamicConv models on WMT 16' En-De dataset with different embedding sharing configurations.}
	\label{DCWMT}
\end{table}	
		
	For Transformer models on IWSLT 14' translation tasks, the 4 new weight sharing methods also outperforms the baseline. Table \ref{TFIWS} shows the results of BLEUs. In this circumstance, square-normalized output embedding method is the best weight sharing method among all methods, which achieves BLEU improvements of 0.86/0.73/0.60 on English to German/Spanish/Romania translation tasks. The Transformer model for IWSLT 14' translation tasks has smaller embedding size. As making the total embedding matrix normalized loses some degrees of freedom for the data of embedding vectors, it will have greater impact on embedding spaces with lower dimensions. Therefore, the method of $l_2$-normalized input embedding can not perform as well as in WMT 16' dataset.
	\begin{table}[H]
		\centering
		\begin{tabular}{lrr}  
			\toprule
			Task  & Configuration & BLEU \\
			\midrule
			De-En        & Baseline  & 34.30      \\
			& $l_2$-normalized input embedding  & 35.00       \\
			& Square-normalized output embedding  &  $\textbf{35.16}$    \\
			& Distances  & 35.09      \\
			& Cosine similarities  & 34.92     \\
			\midrule
			Es-En     & Baseline  & 34.24     \\
			& $l_2$-normalized input embedding  &  34.65  \\
			& Square-normalized output embedding  &  $\textbf{34.97}$    \\
			& Distances  &  34.54      \\
			& Cosine similarities  &  34.81  \\
			\midrule
			Ro-En     & Baseline  &  36.90     \\
			& $l_2$-normalized input embedding  &  37.09  \\
			& Square-normalized output embedding  &  $\textbf{37.50}$     \\
			& Distances  &  37.22     \\
			& Cosine similarities  &  37.34      \\
			\bottomrule
		\end{tabular}
		\caption{BLEUs for Transformer models on IWSLT 14' datasets with different embedding sharing techniques.}
		\label{TFIWS}
	\end{table}
	
	 For DynamicConv models on IWSLT 14' De-En translation tasks, Table \ref{DCIWS} gives the BLEUs of different embedding weight sharing methods. each method has close performance and the method of calculating cosine similarities is a slightly better than others, having a 0.41 BLEU improvement than baseline.

\begin{table}[H]
	\centering
	\begin{tabular}{lrr}  
		\toprule
		Configuration & BLEU \\
		\midrule
		
		Baseline  & 34.52      \\
		l2-normalized input embedding  & 34.86      \\
		Square-normalized output embedding  & 34.74     \\
		Distances  & 34.88      \\
		Cosine similarities  & $\textbf{34.93}$     \\
		
		\bottomrule
	\end{tabular}
	\caption{BLEUs for DynamicConv models on IWSLT 14' En-De dataset (De to En translation) with different embedding sharing techniques.}
	\label{DCIWS}
	
\end{table}	

From all groups of experiments and their results, it can be concluded that the improvements of input-output embedding sharing method work well on state-of-the-art Neural Machine Translation models, and the methods of normalization of input-output shared embedding perform best at most scenraios. The statistical analysis has shown that the methods are unbiased, and the $l_2$-normalized Input Embedding Weight Matrix method even is normalized, which contributes to the effectiveness of them. 

To have a better text generation Neural Network model, normalization of input-output shared embedding is indeed useful. The normalization methods scale the embedding vectors, and make the parts of the Neural Network models more interpretable: The input and output embeddings are mapping between language token spaces and continuous numerical sapces with semantics, the encoder encrypts features with the embedding vectors, and the decoder provides output vectors on the embedding space.

\section{Related Works}
With the fast growth and development of computational power and network training techiques, Neural Networks of great sizes and huge amount of parameters have begun to take advantage and become state-of-the-art models in various tasks in Natural Language Processing. Based on Self-attention Transformer \cite{vaswani2017attention}, BERT \cite{devlin2018bert} and its descendents \cite{liu2019roberta,lan2019albert,raffel2019exploring} dominate multi-task learning in Natural Language Processing. Due to the great number of words and tokens in natural language, each model like these needs to spend a portion of its parameters on building mappings between language words/tokens and numerical tensors, which are also known as input and output embeddings. This need contributes a lot to the computational cost and storage difficulty of the models. 

To deal with this problem, one method is to reduce the structure of embeddings. \cite{raunak2017simple,acharya2019online} present effective methods to make compression of input embeddings, therefore lowers the parameters needed in input embeddings. the shortcoming of them is that this kind of methods is data-based, which means that intensive training on full models is still needed. \cite{lan2019albert} directly factorizes the input embedding layer, and can reduce the parameters of it in any scale. For output mapping, taking the long-tail distribution of words in natural laguages into account, \cite{grave2017efficient} designs a hierarchical Softmax layer for language models, called Adaptive Softmax, which saves a great portion of parameters in the output layer. 

Another kind of parameter reducing techique does not adjust the structure of Neural Networks, but creates parameter sharing patterns in networks. \cite{xia2019tied,lan2019albert} are examples of sharing parameters. \cite{baevski2018adaptive} raised Adaptive Embedding, which is like Adaptive Softmax to make use of different sizes of parameters to mapping word clusters of different frequencies. When Adaptive Embedding and Adaptive Softmax, the parameters can be shared between them in a way.  

In Neural Machine Translation Task, traditional solutionships with Neural Networks are LSTM Network with Attention Mechanism, or Convolutional Networks. Recently, Self-attention based Transformer network shows its advantages: benefited from optimized training schemes and data augmentation techniques such as back translation \cite{edunov2018understanding}, the translation results have outperforming quality. Some improvements and derivations from Transformer have also come out. \cite{dai2019transformer} expands the context width of Transformer to extract more features from context. \cite{wu2019pay} raises new operations: Light Convlution and Dynamic Convolution as substitutions for self-attention in Transformer, and presents competitive test results in the experiments. \cite{gu2019levenshtein} is a Transformer-based model in the non-autoregressive paradigm of Neural Machine Translation.

\section{Conclusion}
To decrease parameters through different parts of Natural Language Processing models, the simple technique of using the same input and output embedding weights matrix is widely adopted by recent state-of-the-art models meanwhile preserves or imporves the performances of text generation models. 

To go further on optimizing its effectiveness, this paper analyzes the reasons of how the embedding sharing technique works, explores its shortcomings, and presents improvements and adjustments of it based on theoretical analysis. By applying normalization methods on embedding weight matrices, the bias of estimations for output scores is eliminated. This paper's work can be applied on various of Neural Networks in Natural Language Processing as long as input-output embedding sharing is included.

In the experiments on various datasets of 2 kinds of Neural Machine Translation models, the improvement methods this paper designs shows guaranteed effectiveness, at the same time nearly do not raise the training and inference time of models at all.

\bibliographystyle{named}
\bibliography{ijcai20}

\begin{thebibliography}{}

\bibitem[\protect\citeauthoryear{Acharya \bgroup \em et al.\egroup
  }{2019}]{acharya2019online}
Anish Acharya, Rahul Goel, Angeliki Metallinou, and Inderjit Dhillon.
\newblock Online embedding compression for text classification using low rank
  matrix factorization.
\newblock In {\em Proceedings of the AAAI Conference on Artificial
  Intelligence}, volume~33, pages 6196--6203, 2019.

\bibitem[\protect\citeauthoryear{Baevski and Auli}{2018}]{baevski2018adaptive}
Alexei Baevski and Michael Auli.
\newblock Adaptive input representations for neural language modeling.
\newblock {\em arXiv preprint arXiv:1809.10853}, 2018.

\bibitem[\protect\citeauthoryear{Dai \bgroup \em et al.\egroup
  }{2019}]{dai2019transformer}
Zihang Dai, Zhilin Yang, Yiming Yang, William~W Cohen, Jaime Carbonell, Quoc~V
  Le, and Ruslan Salakhutdinov.
\newblock Transformer-xl: Attentive language models beyond a fixed-length
  context.
\newblock {\em arXiv preprint arXiv:1901.02860}, 2019.

\bibitem[\protect\citeauthoryear{Devlin \bgroup \em et al.\egroup
  }{2018}]{devlin2018bert}
Jacob Devlin, Ming-Wei Chang, Kenton Lee, and Kristina Toutanova.
\newblock Bert: Pre-training of deep bidirectional transformers for language
  understanding.
\newblock {\em arXiv preprint arXiv:1810.04805}, 2018.

\bibitem[\protect\citeauthoryear{Edunov \bgroup \em et al.\egroup
  }{2017}]{edunov2017classical}
Sergey Edunov, Myle Ott, Michael Auli, David Grangier, and Marc'Aurelio
  Ranzato.
\newblock Classical structured prediction losses for sequence to sequence
  learning.
\newblock {\em arXiv preprint arXiv:1711.04956}, 2017.

\bibitem[\protect\citeauthoryear{Edunov \bgroup \em et al.\egroup
  }{2018}]{edunov2018understanding}
Sergey Edunov, Myle Ott, Michael Auli, and David Grangier.
\newblock Understanding back-translation at scale.
\newblock {\em arXiv preprint arXiv:1808.09381}, 2018.

\bibitem[\protect\citeauthoryear{Grave \bgroup \em et al.\egroup
  }{2017}]{grave2017efficient}
Edouard Grave, Armand Joulin, Moustapha Ciss{\'e}, Herv{\'e} J{\'e}gou, et~al.
\newblock Efficient softmax approximation for gpus.
\newblock In {\em Proceedings of the 34th International Conference on Machine
  Learning-Volume 70}, pages 1302--1310. JMLR. org, 2017.

\bibitem[\protect\citeauthoryear{Gu \bgroup \em et al.\egroup
  }{2019}]{gu2019levenshtein}
Jiatao Gu, Changhan Wang, and Jake Zhao.
\newblock Levenshtein transformer.
\newblock {\em arXiv preprint arXiv:1905.11006}, 2019.

\bibitem[\protect\citeauthoryear{Kudo and
  Richardson}{2018}]{kudo2018sentencepiece}
Taku Kudo and John Richardson.
\newblock Sentencepiece: A simple and language independent subword tokenizer
  and detokenizer for neural text processing.
\newblock {\em arXiv preprint arXiv:1808.06226}, 2018.

\bibitem[\protect\citeauthoryear{Lan \bgroup \em et al.\egroup
  }{2019}]{lan2019albert}
Zhenzhong Lan, Mingda Chen, Sebastian Goodman, Kevin Gimpel, Piyush Sharma, and
  Radu Soricut.
\newblock Albert: A lite bert for self-supervised learning of language
  representations.
\newblock {\em arXiv preprint arXiv:1909.11942}, 2019.

\bibitem[\protect\citeauthoryear{Liu \bgroup \em et al.\egroup
  }{2019}]{liu2019roberta}
Yinhan Liu, Myle Ott, Naman Goyal, Jingfei Du, Mandar Joshi, Danqi Chen, Omer
  Levy, Mike Lewis, Luke Zettlemoyer, and Veselin Stoyanov.
\newblock Roberta: A robustly optimized bert pretraining approach.
\newblock {\em arXiv preprint arXiv:1907.11692}, 2019.

\bibitem[\protect\citeauthoryear{Nguyen and Chiang}{2017}]{nguyen2017improving}
Toan~Q Nguyen and David Chiang.
\newblock Improving lexical choice in neural machine translation.
\newblock {\em arXiv preprint arXiv:1710.01329}, 2017.

\bibitem[\protect\citeauthoryear{Nguyen and
  Salazar}{2019}]{nguyen2019transformers}
Toan~Q Nguyen and Julian Salazar.
\newblock Transformers without tears: Improving the normalization of
  self-attention.
\newblock {\em arXiv preprint arXiv:1910.05895}, 2019.

\bibitem[\protect\citeauthoryear{Ott \bgroup \em et al.\egroup
  }{2018}]{ott2018scaling}
Myle Ott, Sergey Edunov, David Grangier, and Michael Auli.
\newblock Scaling neural machine translation.
\newblock {\em arXiv preprint arXiv:1806.00187}, 2018.

\bibitem[\protect\citeauthoryear{Press and Wolf}{2016}]{press2016using}
Ofir Press and Lior Wolf.
\newblock Using the output embedding to improve language models.
\newblock {\em arXiv preprint arXiv:1608.05859}, 2016.

\bibitem[\protect\citeauthoryear{Raffel \bgroup \em et al.\egroup
  }{2019}]{raffel2019exploring}
Colin Raffel, Noam Shazeer, Adam Roberts, Katherine Lee, Sharan Narang, Michael
  Matena, Yanqi Zhou, Wei Li, and Peter~J Liu.
\newblock Exploring the limits of transfer learning with a unified text-to-text
  transformer.
\newblock {\em arXiv preprint arXiv:1910.10683}, 2019.

\bibitem[\protect\citeauthoryear{Raunak}{2017}]{raunak2017simple}
Vikas Raunak.
\newblock Simple and effective dimensionality reduction for word embeddings.
\newblock {\em arXiv preprint arXiv:1708.03629}, 2017.

\bibitem[\protect\citeauthoryear{Sennrich \bgroup \em et al.\egroup
  }{2015}]{sennrich2015BPE}
Rico Sennrich, Barry Haddow, and Alexandra Birch.
\newblock Neural machine translation of rare words with subword units.
\newblock {\em arXiv preprint arXiv:1508.07909}, 2015.

\bibitem[\protect\citeauthoryear{Sergey \bgroup \em et al.\egroup
  }{2017}]{fairseq}
Edunov Sergey, Ott Myle, and Gross Sam.
\newblock Fairseq.
\newblock https://github.com/pytorch/fairseq, 2017.

\bibitem[\protect\citeauthoryear{Vaswani \bgroup \em et al.\egroup
  }{2017}]{vaswani2017attention}
Ashish Vaswani, Noam Shazeer, Niki Parmar, Jakob Uszkoreit, Llion Jones,
  Aidan~N Gomez, {\L}ukasz Kaiser, and Illia Polosukhin.
\newblock Attention is all you need.
\newblock In {\em Advances in neural information processing systems}, pages
  5998--6008, 2017.

\bibitem[\protect\citeauthoryear{Wu \bgroup \em et al.\egroup
  }{2019}]{wu2019pay}
Felix Wu, Angela Fan, Alexei Baevski, Yann~N Dauphin, and Michael Auli.
\newblock Pay less attention with lightweight and dynamic convolutions.
\newblock {\em arXiv preprint arXiv:1901.10430}, 2019.

\bibitem[\protect\citeauthoryear{Xia \bgroup \em et al.\egroup
  }{2019}]{xia2019tied}
Yingce Xia, Tianyu He, Xu~Tan, Fei Tian, Di~He, and Tao Qin.
\newblock Tied transformers: Neural machine translation with shared encoder and
  decoder.
\newblock In {\em Proceedings of the AAAI Conference on Artificial
  Intelligence}, volume~33, pages 5466--5473, 2019.

\bibitem[\protect\citeauthoryear{Xing \bgroup \em et al.\egroup
  }{2015}]{xing2015normalized}
Chao Xing, Dong Wang, Chao Liu, and Yiye Lin.
\newblock Normalized word embedding and orthogonal transform for bilingual word
  translation.
\newblock In {\em Proceedings of the 2015 Conference of the North American
  Chapter of the Association for Computational Linguistics: Human Language
  Technologies}, pages 1006--1011, 2015.

\bibitem[\protect\citeauthoryear{Yang \bgroup \em et al.\egroup
  }{2019}]{yang2019xlnet}
Zhilin Yang, Zihang Dai, Yiming Yang, Jaime Carbonell, Ruslan Salakhutdinov,
  and Quoc~V Le.
\newblock Xlnet: Generalized autoregressive pretraining for language
  understanding.
\newblock {\em arXiv preprint arXiv:1906.08237}, 2019.

\end{thebibliography}

\end{document}